\documentclass[11pt]{article}

\usepackage[preprint]{acl}

\usepackage{times}
\usepackage{latexsym}
\usepackage[T1]{fontenc}
\usepackage[utf8]{inputenc}
\usepackage{microtype}
\usepackage{inconsolata}

\usepackage{amsmath}
\usepackage{amssymb}
\usepackage{graphicx}
\usepackage{booktabs}
\usepackage{fontawesome}
\usepackage{tikz}
\usetikzlibrary{arrows.meta,positioning,fit,backgrounds,calc}
\usepackage{pgfplots}
\pgfplotsset{compat=1.17}

\definecolor{envcol}{HTML}{2F6F8F}   % environment layer  (teal-blue)
\definecolor{rolcol}{HTML}{B5651D}   % rollout layer      (ochre)
\definecolor{traincol}{HTML}{5B7553} % training layer     (muted green)
\definecolor{inkcol}{HTML}{2B2B2B}   % text / arrows
\definecolor{bandbg}{HTML}{F4F1EC}   % layer band background
\definecolor{rewardcol}{HTML}{1F6FB2}  % reward / success curve (blue)
\definecolor{turncol}{HTML}{C1662A}    % avg-turns curve (orange)

\title{MCP-Universe RL: A Framework for Training MCP Tool-Use Agents via Reinforcement Learning}

\author{
  \textbf{Ziyang Luo} \quad \textbf{Yan Yang} \quad \textbf{Xiangru Jian} \quad \textbf{Ziji Shi} \\
  \textbf{Xiaoqiang Lin} \quad \textbf{Jun Hao Liew} \quad \textbf{Silvio Savarese} \quad \textbf{Junnan Li} \\[2pt]
  Salesforce AI Research\\[3pt]
  \faGlobe~\href{https://mcp-universe.github.io/mcpu-rl}{mcp-universe.github.io/mcpu-rl} \\[3pt]
  \faGithub~\href{https://github.com/SalesforceAIResearch/MCP-Universe/tree/main/mcpuniverse/rl}{SalesforceAIResearch/MCP-Universe}
}

\begin{document}

\maketitle
\begin{abstract}
Reinforcement learning (RL) has become an effective way to improve the tool-use
ability of large language models (LLMs), but most existing RL frameworks stop at the
policy update. For every new domain, the user is left with two hard systems
problems: standing up an isolated environment for each of hundreds of concurrent
trajectories and connecting it to training, and scheduling the rollout so that the
GPU stays busy across long, multi-turn episodes that spend much of their time
stalled on slow tool calls. We present \textbf{MCP-Universe RL (MCP-U RL)}, an
open-source framework that takes over both. It uses the \emph{Model
Context Protocol} (MCP) as the interface to the environment, so any tool already
exposed as an MCP server plugs into training with no RL-specific integration code.
It builds the two missing layers once and reuses them across domains: an
environment-orchestration layer that provisions, isolates, and recycles the MCP
environments over a pluggable container backend, and a rollout-orchestration layer
whose staged pipeline overlaps trajectories to keep the GPU busy while episodes wait
on tools. A backend-agnostic training layer then applies the update through an
existing RL backend, with veRL and slime integrations. With one configuration, changing only the task specification, we train software-engineering, deep-research, and general tool-use agents on \texttt{gpt-oss-20b} and improve task reward in all three.
\end{abstract}
\section{Introduction}
\label{sec:intro}

Large language models (LLMs) are increasingly expected to act in the real world through external tools~\cite{luo2025mcpuniverse}, such as running code, sending emails, querying databases, or operating real applications, rather than producing text answers alone. Reinforcement learning (RL) has become an effective way to improve this behavior: by rewarding task completion, algorithms such as GRPO~\cite{shao2024deepseekmath,deepseekai2025r1} and DAPO~\cite{yu2025dapo} teach an agent to plan, call tools, and recover from errors over many turns. Putting this into practice takes more than the algorithm. General RL frameworks~\cite{sheng2024hybridflow,hu2024openrlhf,thudm2025slime,areal2025} provide the optimization, the policy update and distributed training, but to train a tool-use agent the user still has to build two further layers for every new domain.

The first is the \emph{environment-orchestration} layer. Agentic RL samples many trajectories per
update, and each one acts on a live environment whose state it changes, editing
files, writing database rows, or navigating a stateful session~\cite{Jimenez2023SWEbenchCL,Chezelles2024TheBE}. The reward is usually read
back from that state, so if two trajectories shared an environment their edits would
collide and the reward would be wrong; each trajectory therefore needs its own
isolated environment. At the scale of RL, hundreds of these are started on a
container backend, kept reachable, and recycled at once, which is a systems problem
in itself. Yet the work is nearly the same from one domain to the next: a
web-search environment and a code repository differ in what they run, not in how
they are provisioned and recycled. The second is the
\emph{rollout-orchestration} layer that turns those environments into training signal. Each
tool-use turn alternates between generating on the GPU and calling a tool that runs
in the environment, and a tool call can stall for seconds. Running one trajectory at
a time leaves the GPU idle through every such stall; running many at once keeps it
busy but can exhaust host memory with live environments. Keeping the GPU fed without
running out of memory is therefore a scheduling problem. Both layers are rebuilt per
project today, which is what makes agentic RL slow to set up and hard to reproduce.

\begin{figure*}[t]
  \centering
  \includegraphics[width=0.98\textwidth]{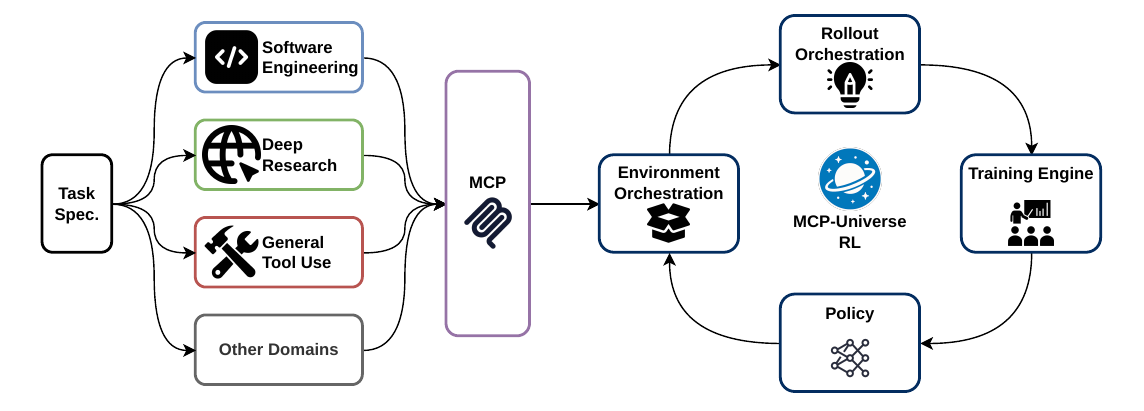}
  \caption{Overview of our MCP-Universe RL framework. Each domain is exposed as one or
  several MCP servers, and all domains funnel through a single MCP interface into one
  shared training loop, whose environment-orchestration, rollout-orchestration, and
  training-engine layers are reused for every domain; only the task specification
  changes.}
  \label{fig:overview}
\end{figure*}

In this work, we present \textbf{MCP-Universe RL (MCP-U RL)}, an open-source framework that builds
both layers once and shares them across domains (Figure~\ref{fig:overview}). The key
idea is to use the \emph{Model Context Protocol} (MCP)~\cite{anthropic2024mcp} as a
single interface to every environment: MCP standardizes how an agent calls tools, and
a large ecosystem of MCP servers already exists. Once a domain's tools are exposed as
an MCP server, the user only writes a task specification and presses run, with no
environment-integration or rollout code. Because every domain is now reached the same
way, the \emph{environment-orchestration} layer manages the whole life cycle in one
place, provisioning, isolating, and recycling environments over a container backend
that can be swapped when one runtime cannot create them fast enough.

Each ready environment is then handed to the \emph{rollout-orchestration} layer, which
runs the episodes and turns them into training signal: for each trajectory it drives
the agent against its environment, records the tokens for training, and scores the
finished episode with the task's evaluator. Running these episodes fast is the hard part, because a tool-use episode leaves the
GPU idle whenever the agent is waiting on a tool. The layer hides that wait by
running the episodes as an overlapping pipeline, so that some trajectories generate
on the GPU while others are acquiring an environment or waiting on a tool call.
Acquiring an environment and generating draw on different resources, host memory
and the GPU, so we let more trajectories generate than acquire, which keeps the GPU
busy without running out of memory. A backend-agnostic \emph{training} layer then applies the
policy update through an existing RL backend, with integrations for
veRL~\cite{sheng2024hybridflow} and slime~\cite{thudm2025slime}. Changing only the
task specification, we train software-engineering~\cite{r2egym2025}, deep-research~\cite{deepdive2025}, and general
tool-use~\cite{awm2026} agents, and improve task reward in all three.

\section{Related Work}
\label{sec:related}

\begin{figure*}[t!]
  \centering
  \includegraphics[width=\textwidth]{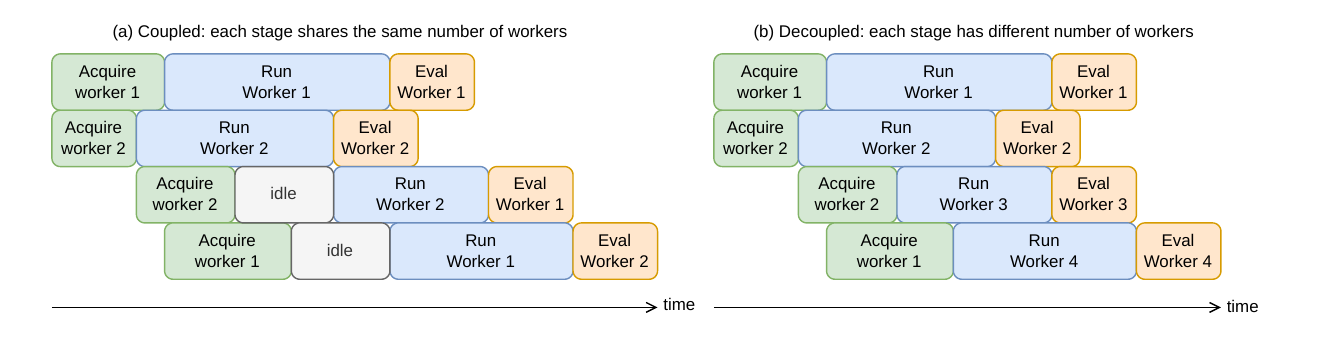}
  \caption{Per-stage concurrency in the rollout pipeline. Each row is one trajectory,
  flowing Acquire\,$\rightarrow$\,Run\,$\rightarrow$\,Eval left to right; block width is
  relative wall-clock. \textbf{(a)}~Coupled: when every stage is given the same number
  of workers, a trajectory that finishes Acquire must wait for a free Run worker,
  leaving idle gaps. \textbf{(b)}~Decoupled: giving the GPU-bound Run stage more workers than the fast Acquire stage lets trajectories flow through with less
  idle.}
  \label{fig:loop}
\end{figure*}

General RL frameworks such as veRL~\cite{sheng2024hybridflow}, OpenRLHF~\cite{hu2024openrlhf}, slime~\cite{thudm2025slime}, and AReaL~\cite{areal2025}
supply the optimization backend but leave the environment, the reward, and the
multi-turn rollout to the user. MCP-U RL integrates veRL and slime as interchangeable
backends and exposes the same adapter interface for adding others, while supplying
the missing environment, reward, and rollout layers. A second line of work builds the
agentic rollout on top of these frameworks: SkyRL-Agent~\cite{skyrlagent2025} runs a staged asynchronous pipeline,
AgentRL~\cite{agentrl2025} a multi-turn multi-task framework,
VerlTool~\cite{verltool2025} a standardized tool API, and rLLM~\cite{rllm2025} many
agent harnesses in pluggable sandboxes. All still ask the user to wire each tool or
harness in behind their own interface, so a new domain means new integration code;
MCP-U RL instead makes MCP the single interface, so any existing MCP server plugs
into training unchanged while the environment-orchestration layer manages its life
cycle, and our rollout engine exposes its scheduling as configuration.

The work closest to ours shares either our claim or our interface. Agent
Lightning~\cite{agentlightning2025} traces a running agent so that almost any agent
framework can be optimized without code change; it makes the \emph{agent} pluggable
but still leaves the tools and their environments orchestration to build, the layer MCP-U RL
supplies. A few efforts instead train agents over MCP tools directly. OpenPipe's
MCP-RL~\cite{openpipe2025mcprl} trains a model to use a single MCP server's tools,
released as software with a demo rather than a cross-domain study.
MiroRL~\cite{mirorl2025} targets the single deep-research domain, and
MCP-Flow~\cite{mcpflow2025} scales MCP tool use through supervised fine-tuning rather
than RL. Each of these targets a single server or domain; MCP-U RL instead trains
across domains from one framework, with rewards from explicit evaluators run against
the environment state.
\section{MCP-Universe RL Framework}\label{sec:design}

MCP-U RL has three layers, split by the resource each one uses
(Figure~\ref{fig:overview}). The environment-orchestration layer runs the MCP
environments on the host, the rollout-orchestration layer runs the episodes between
the host and the GPU, and the training-engine layer applies the policy update on the
GPU. The layers pass data, not domain logic, so the only thing that changes from one
domain to the next is the task specification the user writes.

\subsection{Environment-Orchestration Layer}
\label{sec:design:env}

Agentic RL samples many trajectories at once, and each acts on an environment whose
state it changes, editing files, writing database rows, or navigating a stateful web
session. Sharing an environment would let these changes collide and corrupt the
reward, which is read from the final state, so each trajectory needs its own
isolated environment. Standing that up, keeping it reachable, and tearing it down is
real work that existing frameworks leave to the user; we make it a layer with an
interface on each side, one hiding what the tools are and one hiding how they are
run. On the agent's side, the environment's tools are served behind a
per-environment MCP gateway, so an environment, whether it wraps a browser, a code
repository, or a database, is reached the same way: an address the rollout layer
calls over MCP. On the runtime side, a \emph{provisioner} exposes \texttt{create},
\texttt{reset}, \texttt{health\_check}, and \texttt{destroy}, so no specific runtime
(e.g., Docker) is hard-coded anywhere else.

Between the two interfaces, the layer manages the life cycle of each environment. It
acquires an environment, runs an optional \texttt{prepares} step for per-instance
setup, and holds it through both the episode and its evaluation, since the reward is
often read from the live environment, for example when a hidden test suite runs in
the same container the agent just edited. It then resets or destroys the environment.
To avoid building one environment per trajectory, the layer keeps a pool. When a
domain's tasks share a configuration, the pool is pre-warmed and released
environments are reset and reused. When each task needs a distinct image, as in
software engineering where every instance carries its own repository, environments
are provisioned on demand instead. These two cases stress environment creation in
very different ways, and no single runtime handles both well. That is why the
provisioner is an interface: we ship one backend on Docker and one daemonless, and a
new isolation mechanism is added simply by implementing the same interface.

\subsection{Rollout-Orchestration Layer}
\label{sec:design:rollout}

A tool-use episode does not keep the GPU busy on its own: each turn alternates
between generation on the GPU and a tool call that runs in the environment and can
stall for seconds. Running trajectories one at a time idles the GPU through every
tool call, so this layer overlaps many trajectories, turning each episode into a
scored, tokenized trajectory while others generate. It records the trajectory at the
token level, marking which tokens the agent produced and are trainable, and runs the
task's evaluator for the reward. The agent type and prompt format are set by
configuration, so the same code serves different models and agent styles.

The layer is a single engine organized as a three-stage pipeline whose stages
overlap: acquire the environment, run the episode, and evaluate it
(Figure~\ref{fig:loop}), following prior agentic-RL rollout systems~\cite{skyrlagent2025}. Rather than fix one scheduling policy, we expose the engine's
behavior as configuration; the control we rely on most is \emph{per-stage
concurrency}, which sizes the three stages' worker pools separately. This matters
because the stages are bound by different resources. Acquiring an environment is
host-memory and CPU bound, while the run stage is GPU bound (dominated by generation). We
therefore give the run stage more workers than the acquire stage (at least twice as many by
default), which keeps the GPU busy during other trajectories' tool calls without
holding more environments in memory.

Moreover, the same engine also runs under different \emph{placements} of the GPUs relative to
the update, again by configuration. A colocated placement shares GPUs between
rollout and the update and alternates; a fully asynchronous placement runs them on
separate GPUs and streams trajectories through a queue so generation and the update
proceed at once. Both drive the one engine, so switching between them is a
configuration change rather than a reimplementation.

\subsection{Training-Engine Layer}
\label{sec:design:train}

The rollout layer emits its batch in a backend-neutral form, prompt and response
token ids, a trainable mask, and the per-trajectory reward, so the update itself can
be left to an existing RL backend and this layer only adapts the batch to one. We
provide adapters for veRL~\cite{sheng2024hybridflow} and slime~\cite{thudm2025slime}, and a new one is added by writing the adapter. The backend,
the RL algorithm, and the GPU placement (Section~\ref{sec:design:rollout}) are all
configuration and do not change the layers above.

\subsection{Task Specification}
\label{sec:design:task}

Once a domain's tools are available over MCP, the user drives the framework through
a task specification alone (Figure~\ref{fig:taskspec}). It is a JSON object naming
the MCP servers (\texttt{mcp\_servers}), the \texttt{instruction}, and one or more
\texttt{evaluators} whose score becomes the reward; we adopt the format of
MCP-Universe \cite{luo2025mcpuniverse}. For a stateless task these three fields suffice. A stateful domain adds
optional fields, a \texttt{dockerfile\_path} and \texttt{build\_args} for the image
and \texttt{prepares}/\texttt{cleanups} hooks that seed and undo environment state,
and these are the only place per-domain logic lives. Each hook and evaluator is a
named function in a small registry, so extending to a new domain means writing one
such function and referring to it, not changing the layers above.

\begin{figure}[t]
\centering
\small
\begin{verbatim}
{
  "instance_id": "dr_train_0001",
  "instruction": "Seek the title of a 2012
      paper published in an international
      journal focused on engineering
      science ...",
  "mcp_servers": [{"name": "serper-search"},
                  {"name": "jina-scrape"}],
  "evaluators": [{"func": "raw",
      "op": "deepresearch.llm_as_judge",
      "op_args": {"correct_answer":
          "Grains of Saws ..."}}]

  // optional, for a stateful domain (e.g. SWE):
  "dockerfile_path": "Dockerfile.SWE",
  "build_args": {"SWE_BASE_IMAGE": "..."},
  "prepares": [{"prepare_func": "swe_setup"}],
  "cleanups": [{"cleanup_func": "swe_teardown"}]
}
\end{verbatim}
\caption{A task specification. The three required fields (\texttt{mcp\_servers},
\texttt{instruction}, \texttt{evaluators}) define a stateless task; a stateful
domain adds the optional fields below, each referring to a named registry function.}
\label{fig:taskspec}
\end{figure}

\section{Experiments}
\label{sec:exp}

We evaluate MCP-U RL along its two contributions. First, that the framework can
train and improve agents, using one configuration and changing only the task
specification, across domains whose environments and reward checks differ
(Section~\ref{sec:exp:capability}). Second, that the rollout layer's design choices
raise throughput without changing what is trained (Section~\ref{sec:exp:system}).
We do not aim for state of the art on any benchmark; reaching it would take
dedicated data and environment engineering, which is orthogonal to the framework.

Across all runs, the policy is the open-weight \texttt{gpt-oss-20b} model
\cite{openai2025gptoss}, run as a tool-use agent. Training uses GRPO
\cite{shao2024deepseekmath}, with a learning rate of
$2\times10^{-6}$ and $n=8$ trajectories sampled per task. Rewards are
binary and come entirely from each task's evaluator. Full settings are
in Appendix~\ref{sec:appendix:train}.

\subsection{Training Agents Across Domains}
\label{sec:exp:capability}

We train in three domains that differ substantially in their environments, their tools, their tasks,
and how they are graded. The framework code and its configuration are the same
across all three; only the task specification changes, which is the property we are testing.

\begin{figure}[t]
  \centering
  \includegraphics[width=0.85\linewidth]{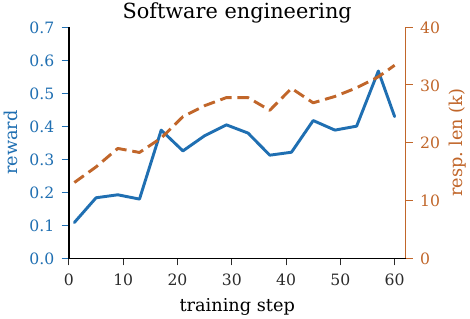}
  \caption{Software-engineering training on R2E-Gym. Left axis (solid, blue):
  reward; right axis (dashed, orange): average response length in
  thousands of tokens.}
  \vspace{-0.4cm}
  \label{fig:curve-swe}
\end{figure}

\begin{figure}[t]
  \centering
  \includegraphics[width=0.85\linewidth]{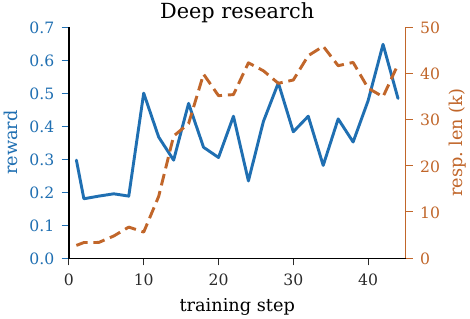}
  \caption{Deep-research training on DeepDive. Axes as in
  Figure~\ref{fig:curve-swe}.}
  \vspace{-0.5cm}
  \label{fig:curve-dr}
\end{figure}

\paragraph{Software engineering} is a bug-fixing task:
given a real bug report on a code repository, the agent must find the faulty code,
edit it, and make the project's tests pass. We train on R2E-Gym~\cite{r2egym2025}, a
collection of real GitHub issues where each task pairs an issue with a container
image of the repository at the buggy commit and a hidden test suite. The environment is stateful:
a \texttt{prepares} step checks out the repository, the agent inspects and edits the
code over many turns, and the evaluator runs the hidden tests inside the same
container for a binary reward on whether they pass. Following
mini-swe-agent~\cite{minisweagent2025}, the agent is given a single shell tool,
exposed over MCP, and nothing else: it reads, searches, edits, and runs tests only by
issuing shell commands. This is the hardest of our three settings, as the agent has
to localize the bug, change the code, and re-run the tests with no specialized
editing or navigation tools, and episodes run the longest as a result. Over 60 steps
of training, the success rate rises from about $0.11$ to about $0.43$ (peaking near
$0.6$), and the mean response length grows from $13$k to $33$k tokens as the agent
learns to take more investigative turns before editing (Figure~\ref{fig:curve-swe}).

\paragraph{Deep research} is a web-browsing task: given a hard question,
the agent must gather and cross-check evidence from many web pages before it can answer,
rather than retrieving a single fact. We train on DeepDive~\cite{deepdive2025}, a set
of hard, multi-hop questions whose answers are not on any one page. The agent runs a
ReAct~\cite{yao2023react} loop, alternating a reasoning step with a tool call, and
reaches the open web through two stateless MCP servers: a Google Search server for
queries and a Jina Scrape server for reading pages. It searches and follows links
over many turns, then commits to a final answer that the evaluator scores against the
reference. Unlike the software-engineering setting there is no container state to
manage, only external web tools, yet the rollout and training code are unchanged.
Over training the success rate rises from about $0.22$ to about $0.52$ (peaking at
$0.65$), and the mean response length climbs more than tenfold, from under $3$k to
around $40$k tokens, as the agent learns to search and read across more pages before
answering rather than replying from the first hit (Figure~\ref{fig:curve-dr}).

\begin{figure}[t]
  \centering
  \includegraphics[width=0.85\linewidth]{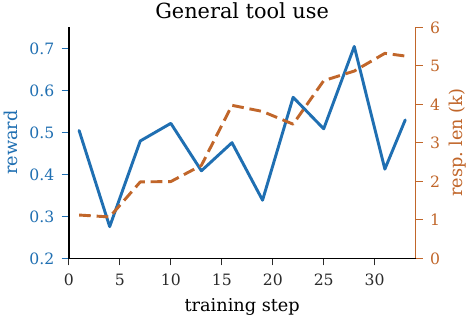}
  \caption{General-tool-use training on AgentWorldModel-1k. Axes as in
  Figure~\ref{fig:curve-swe}.}
  \vspace{-0.5cm}
  \label{fig:curve-tool}
\end{figure}

\paragraph{General tool use} is a broad, multi-application setting: given an everyday
instruction, such as placing an order on a shopping site, moving money between
accounts, or updating a customer record, the agent must drive one or more
applications through their tools to carry the task out. We train on the synthetic environments of AgentWorldModel~\cite{awm2026}, which span everyday
domains including e-commerce, finance, travel, social media, and customer-management
(CRM) applications. Each environment is code- and database-backed, so the agent acts on real
state over many turns; each task gives an instruction whose completion changes
application state, and the evaluator grades success by querying that state after the
episode. We expose each environment's tools over MCP, so the toolset and the checked
state differ again from the previous two domains, spanning many applications rather
than one shell or the web, yet no framework code changes. The policy already starts
around $0.48$ here, and over training the success rate rises modestly to about $0.55$
(peaking near $0.7$); more strikingly, the mean response length grows fivefold, from
$1$k to over $5$k tokens, as the agent stops answering from the surface and learns to
probe each application's state and chain many tool calls before it commits
(Figure~\ref{fig:curve-tool}).

\subsection{Rollout Throughput}
\label{sec:exp:system}

The rollout layer is one engine whose behavior is set by configuration
(Section~\ref{sec:design:rollout}), and the configuration changes how fast
trajectories are produced, not what is produced: every setting samples from the same
policy and scores with the same evaluator, so the training signal is identical and
only the wall-clock differs. We measure throughput on the software-engineering
workload, whose long, tool-heavy episodes stress the rollout layer the most.

\begin{figure}[t]
  \centering
  \includegraphics[width=0.85\linewidth]{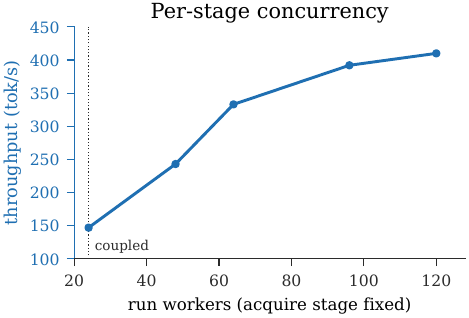}
  \caption{Per-stage concurrency. Rollout throughput
  as the number of run workers grows, with the acquire stage and live-environment
  count fixed. The dotted line marks the coupled point (run = acquire); adding run
  workers is the decoupled setting.}
  \label{fig:decouple}
\end{figure}

The control that matters most is per-stage concurrency
(Section~\ref{sec:design:rollout}). Acquiring an environment is host-memory bound
while the run stage is GPU bound (dominated by generation), so we hold the acquire
stage fixed and vary the number of run workers, keeping the number of live
environments (and thus the memory budget) constant throughout
(Figure~\ref{fig:decouple}). The point where run and acquire share one limit is the
\emph{coupled} setting; there the GPU sits idle whenever that limit is spent on
trajectories still acquiring an environment or waiting on a tool. Adding run workers,
our \emph{decoupled} setting, lets already-acquired trajectories keep running:
rollout throughput rises from $147$ to $410$ tokens per second, a $2.8\times$ gain,
and then saturates as generation fills the GPU, all without holding more environments in memory.

The same holds across GPU \emph{placements}. Table~\ref{tab:placement} reports
end-to-end training throughput under the two at matched concurrency: the colocated
placement shares GPUs between rollout and the update so they wait on each other,
while the fully asynchronous placement runs them on separate GPUs at once and is
about $2\times$ faster. Our asynchronous placement is one-step off-policy, updating
on trajectories at most one step stale, so it stays close to on-policy while still
overlapping rollout and update. Full settings for all rollout experiments are in
Appendix~\ref{sec:appendix:rollout}.

\begin{table}[t]
  \centering
  \small
  \resizebox{\columnwidth}{!}{%
  \begin{tabular}{lcc}
    \toprule
    GPU placement & Time / step (s) & Throughput (steps/h) \\
    \midrule
    Colocated                 & $1164$          & $3.1$ \\
    Fully asynchronous        & $\mathbf{572}$  & $\mathbf{6.3}$ \\
    \bottomrule
  \end{tabular}}
  \caption{End-to-end training throughput under the two GPU placements, same engine
  and workload.}
  \vspace{-0.5cm}
  \label{tab:placement}
\end{table}

\section{Conclusion}
\label{sec:conclusion}

We presented MCP-U RL, a framework that trains tool-use agents with RL by using MCP as the interface to the environment. It builds the environment-orchestration and rollout-orchestration layers once and reuses them across domains, so a new domain is added by writing a task specification rather than RL integration code. With the same framework and only the task specification changed, we trained software-engineering, deep-research, and general-tool-use agents and improved task reward in all three; decoupling the rollout stages further raised throughput by $2.8\times$ without altering the training signal. The framework is open source, and we hope it lowers the cost of training agents over the growing MCP ecosystem.

% Bibliography (acl.sty already sets \bibliographystyle{acl_natbib})
\bibliography{references}

@misc{anthropic2024mcp,
  title={Introducing the {Model Context Protocol}},
  author={{Anthropic}},
  year={2024},
  howpublished={\url{https://www.anthropic.com/news/model-context-protocol}}
}

@article{luo2025mcpuniverse,
  title={MCP-Universe: Benchmarking Large Language Models with Real-World Model Context Protocol Servers},
  author={Ziyang Luo and Zhiqi Shen and Wenzhuo Yang and Zirui Zhao and Prathyusha Jwalapuram and Amrita Saha and Doyen Sahoo and Silvio Savarese and Caiming Xiong and Junnan Li},
  journal={ArXiv},
  year={2025},
  volume={abs/2508.14704},
  url={https://api.semanticscholar.org/CorpusID:280691759}
}

@article{yao2023react,
  title={ReAct: Synergizing Reasoning and Acting in Language Models},
  author={Shunyu Yao and Jeffrey Zhao and Dian Yu and Nan Du and Izhak Shafran and Karthik Narasimhan and Yuan Cao},
  journal={ArXiv},
  year={2022},
  volume={abs/2210.03629},
  url={https://api.semanticscholar.org/CorpusID:252762395}
}

@article{shao2024deepseekmath,
  title={DeepSeekMath: Pushing the Limits of Mathematical Reasoning in Open Language Models},
  author={Zhihong Shao and Peiyi Wang and Qihao Zhu and Runxin Xu and Jun-Mei Song and Mingchuan Zhang and Y. K. Li and Yu Wu and Daya Guo},
  journal={ArXiv},
  year={2024},
  volume={abs/2402.03300},
  url={https://api.semanticscholar.org/CorpusID:267412607}
}

@article{yu2025dapo,
  title={DAPO: An Open-Source LLM Reinforcement Learning System at Scale},
  author={Qiying Yu and Zheng Zhang and Ruofei Zhu and Yufeng Yuan and Xiaochen Zuo and Yu Yue and Tiantian Fan and Gaohong Liu and Lingjun Liu and Xin Liu and Haibin Lin and Zhiqi Lin and Bole Ma and Guangming Sheng and Yuxuan Tong and Chi Zhang and Mofan Zhang and Wang Zhang and Hang Zhu and Jinhua Zhu and Jiaze Chen and Jiangjie Chen and Chengyi Wang and Honglin Yu and Weinan Dai and Yuxuan Song and Xiang Wei and Haodong Zhou and Jingjing Liu and Wei Ma and Ya-Qin Zhang and Lin Yan and Mu Qiao and Yong-Xu Wu and Mingxuan Wang},
  journal={ArXiv},
  year={2025},
  volume={abs/2503.14476},
  url={https://api.semanticscholar.org/CorpusID:277104124}
}

@article{deepseekai2025r1,
  title={DeepSeek-R1 incentivizes reasoning in LLMs through reinforcement learning},
  author={DeepSeek-AI and Daya Guo and Dejian Yang and Haowei Zhang and Jun-Mei Song and Ruoyu Zhang and Runxin Xu and Qihao Zhu and Shirong Ma and Peiyi Wang and Xiaoling Bi and Xiaokang Zhang and Xingkai Yu and Yu Wu and Z. F. Wu and Zhibin Gou and Zhihong Shao and Zhuoshu Li and Ziyi Gao and Aixin Liu and Bing Xue and Bing-Li Wang and Bochao Wu and Bei Feng and Chengda Lu and Chenggang Zhao and Chengqi Deng and Chenyu Zhang and Chong Ruan and Damai Dai and Deli Chen and Dong-Li Ji and Erhang Li and Fangyun Lin and Fucong Dai and Fuli Luo and Guangbo Hao and Guanting Chen and Guowei Li and H. Zhang and Han Bao and Hanwei Xu and Haocheng Wang and Honghui Ding and Huajian Xin and Huazuo Gao and Hui Qu and Hui Li and Jianzhong Guo and Jiashi Li and Jiawei Wang and JingChang Chen and Jingyang Yuan and Junjie Qiu and Junlong Li and J. L. Cai and Jiaqi Ni and Jian Liang and Jin Chen and Kai Dong and Kai Hu and Kaige Gao and Kang Guan and Kexin Huang and Kuai Yu and Lean Wang and Lecong Zhang and Liang Zhao and Litong Wang and Liyue Zhang and Lei Xu and Leyi Xia and Mingchuan Zhang and Minghua Zhang and M. Tang and Meng Li and Miaojun Wang and Mingming Li and Ning Tian and Panpan Huang and Peng Zhang and Qiancheng Wang and Qinyu Chen and Qiushi Du and Ruiqi Ge and Ruisong Zhang and Ruizhe Pan and Runji Wang and R. J. Chen and Rui-Qi Jin and Ruyi Chen and Shanghao Lu and Shangyan Zhou and Shanhuang Chen and Shengfeng Ye and Shiyu Wang and Shuiping Yu and Shunfeng Zhou and Shuting Pan and S. S. Li and Shuang Zhou and Shao-Kang Wu and Tao Yun and Tian Pei and T. Sun and T. Wang and Wangding Zeng and Wanjia Zhao and Wen Liu and Wenfeng Liang and Wenjun Gao and Wen-Xia Yu and Wentao Zhang and Wangding Xiao and Wei An and Xiaodong Liu and Xiaohan Wang and Xiaokang Chen and Xiaotao Nie and Xin Cheng and Xin Liu and Xin Xie and Xingchao Liu and Xinyu Yang and Xinyuan Li and Xuecheng Su and Xuheng Lin and X. Q. Li and Xiangyu Jin and Xi-Cheng Shen and Xiaosha Chen and Xiaowen Sun and Xiaoxiang Wang and Xinnan Song and Xinyi Zhou and Xianzu Wang and Xinxia Shan and Y. K. Li and Y. Q. Wang and Y. X. Wei and Yang Zhang and Yanhong Xu and Yao Li and Yao Zhao and Yaofeng Sun and Yaohui Wang and Yi Yu and Yichao Zhang and Yifan Shi and Yi Xiong and Ying He and Yishi Piao and Yisong Wang and Yixuan Tan and Yiyang Ma and Yiyuan Liu and Yongqiang Guo and Yuan Ou and Yuduan Wang and Yue Gong and Yu-Jing Zou and Yujia He and Yunfan Xiong and Yu-Wei Luo and Yu-mei You and Yuxuan Liu and Yuyang Zhou and Y. X. Zhu and Yanping Huang and Yao Li and Yi Zheng and Yuchen Zhu and Yunxiang Ma and Ying Tang and Yukun Zha and Yuting Yan and Zehui Ren and Zehui Ren and Zhangli Sha and Zhe Fu and Zhean Xu and Zhenda Xie and Zhen-guo Zhang and Zhewen Hao and Zhicheng Ma and Zhigang Yan and Zhiyu Wu and Zihui Gu and Zijia Zhu and Zijun Liu and Zi-Long Li and Ziwei Xie and Ziyang Song and Zizheng Pan and Zhen Huang and Zhipeng Xu and Zhongyu Zhang and Zhen Zhang},
  journal={Nature},
  year={2025},
  volume={645},
  pages={633 - 638},
  url={https://api.semanticscholar.org/CorpusID:275789950}
}

@article{skyrlagent2025,
  title={SkyRL-Agent: Efficient RL Training for Multi-turn LLM Agent},
  author={Shiyi Cao and Dacheng Li and Fangzhou Zhao and Shuo Yuan and Sumanth Hegde and Connor Chen and Charlie Ruan and Tyler Griggs and Shu Liu and Eric Tang and Richard Liaw and Philipp Moritz and Matei Zaharia and Joseph Gonzalez and Ion Stoica},
  journal={ArXiv},
  year={2025},
  volume={abs/2511.16108},
  url={https://api.semanticscholar.org/CorpusID:283110250}
}

@article{sheng2024hybridflow,
  title={HybridFlow: A Flexible and Efficient RLHF Framework},
  author={Guangming Sheng and Chi Zhang and Zilingfeng Ye and Xibin Wu and Wang Zhang and Ru Zhang and Yanghua Peng and Haibin Lin and Chuan Wu},
  journal={Proceedings of the Twentieth European Conference on Computer Systems},
  year={2025},
  url={https://api.semanticscholar.org/CorpusID:272987758}
}

@article{hu2024openrlhf,
  title={OpenRLHF: An Easy-to-use, Scalable and High-performance RLHF Framework},
  author={Jian Hu and Xibin Wu and Weixun Wang and Songlin Jiang and Dehao Zhang and Yu Cao and OpenLLMAI Team and Netease Fuxi and AI Lab and Alibaba Group},
  journal={ArXiv},
  year={2024},
  volume={abs/2405.11143},
  url={https://api.semanticscholar.org/CorpusID:269921667}
}

@misc{thudm2025slime,
  title={slime: An {LLM} post-training framework for {RL} scaling},
  author={{THUDM}},
  year={2025},
  howpublished={\url{https://github.com/THUDM/slime}}
}

@article{areal2025,
  title={AReaL: A Large-Scale Asynchronous Reinforcement Learning System for Language Reasoning},
  author={Wei Fu and Jiaxuan Gao and Xu Shen and Chen Zhu and Zhiyu Mei and Chuyi He and Shusheng Xu and Guoyizhe Wei and Jun Mei and Jiashun Wang and Tongkai Yang and Binhang Yuan and Yi Wu},
  journal={ArXiv},
  year={2025},
  volume={abs/2505.24298},
  url={https://api.semanticscholar.org/CorpusID:279071008}
}

@article{r2egym2025,
  title={R2E-Gym: Procedural Environments and Hybrid Verifiers for Scaling Open-Weights SWE Agents},
  author={Naman Jain and Jaskirat Singh and Manish Shetty and Liang Zheng and Koushik Sen and Ion Stoica},
  journal={ArXiv},
  year={2025},
  volume={abs/2504.07164},
  url={https://api.semanticscholar.org/CorpusID:277667306}
}

@misc{minisweagent2025,
  title={{mini-swe-agent}: The 100-line {AI} agent that solves {GitHub} issues},
  author={{SWE-agent Team}},
  year={2025},
  howpublished={\url{https://github.com/SWE-agent/mini-swe-agent}}
}

@article{deepdive2025,
  title={DeepDive: Advancing Deep Search Agents with Knowledge Graphs and Multi-Turn RL},
  author={Rui Lu and Zhenyu Hou and Zihan Wang and Hanchen Zhang and Xiao Liu and Yujiang Li and Shi Feng and Jie Tang and Yuxiao Dong},
  journal={ArXiv},
  year={2025},
  volume={abs/2509.10446},
  url={https://api.semanticscholar.org/CorpusID:281310021}
}

@misc{openai2025gptoss,
  title={{gpt-oss-120b} and {gpt-oss-20b} Model Card},
  author={{OpenAI}},
  year={2025},
  howpublished={\url{https://openai.com/index/introducing-gpt-oss/}}
}

@misc{openpipe2025mcprl,
  title={{MCP-RL}: Teach any model to master any {MCP} server},
  author={{OpenPipe}},
  year={2025},
  howpublished={\url{https://art.openpipe.ai/features/mcp-rl}}
}

@misc{mirorl2025,
  title={{MiroRL}: An {MCP}-first Reinforcement Learning Framework for Deep Research Agents},
  author={{MiroMind AI}},
  year={2025},
  howpublished={\url{https://github.com/MiroMindAI/MiroRL}}
}

@article{mcpflow2025,
  title={MCP-Flow: Facilitating LLM Agents to Master Real-World, Diverse and Scaling MCP Tools},
  author={Wenhao Wang and Peizhi Niu and Zhao Xu and Zhaoyu Chen and Jian Du and Yaxin Du and Xianghe Pang and Keduan Huang and Yanfeng Wang and Qiang Yan and Siheng Chen},
  journal={ArXiv},
  year={2025},
  volume={abs/2510.24284},
  url={https://api.semanticscholar.org/CorpusID:282400958}
}

@article{agentrl2025,
  title={AgentRL: Scaling Agentic Reinforcement Learning with a Multi-Turn, Multi-Task Framework},
  author={Hanchen Zhang and Xiao Liu and Bowen Lv and Xueqiao Sun and Bohao Jing and Iat Long Iong and Zhenyu Hou and Zehan Qi and Hanyu Lai and Yifan Xu and Rui Lu and Hongning Wang and Jie Tang and Yuxiao Dong},
  journal={ArXiv},
  year={2025},
  volume={abs/2510.04206},
  url={https://api.semanticscholar.org/CorpusID:281842672}
}

@article{Chezelles2024TheBE,
  title={The BrowserGym Ecosystem for Web Agent Research},
  author={Thibault Le Sellier de Chezelles and Maxime Gasse and Alexandre Lacoste and Alexandre Drouin and Massimo Caccia and L'eo Boisvert and Megh Thakkar and Tom Marty and Rim Assouel and Sahar Omidi Shayegan and Lawrence Keunho Jang and Xing Han L{\`u} and Ori Yoran and Dehan Kong and Frank F. Xu and Siva Reddy and Quentin Cappart and Graham Neubig and Ruslan Salakhutdinov and Nicolas Chapados},
  journal={ArXiv},
  year={2024},
  volume={abs/2412.05467},
  url={https://api.semanticscholar.org/CorpusID:274598279}
}

@article{Jimenez2023SWEbenchCL,
  title={SWE-bench: Can Language Models Resolve Real-World GitHub Issues?},
  author={Carlos E. Jimenez and John Yang and Alexander Wettig and Shunyu Yao and Kexin Pei and Ofir Press and Karthik Narasimhan},
  journal={ArXiv},
  year={2023},
  volume={abs/2310.06770},
  url={https://api.semanticscholar.org/CorpusID:263829697}
}

@article{awm2026,
  title={Agent World Model: Infinity Synthetic Environments for Agentic Reinforcement Learning},
  author={Zhaoyang Wang and Canwen Xu and Boyi Liu and Yite Wang and Siwei Han and Zhewei Yao and Huaxiu Yao and Yuxiong He},
  journal={arXiv preprint arXiv:2602.10090},
  year={2026},
  note={Accepted to ICML 2026}
}

@article{verltool2025,
  title={VerlTool: Towards Holistic Agentic Reinforcement Learning with Tool Use},
  author={Dongfu Jiang and Yi Lu and Zhuofeng Li and Zhiheng Lyu and Ping Nie and Haozhe Wang and Alex Su and Hui Chen and Kai Zou and Chao Du and Tianyu Pang and Wenhu Chen},
  journal={ArXiv},
  year={2025},
  volume={abs/2509.01055},
  url={https://api.semanticscholar.org/CorpusID:281080546}
}

@misc{rllm2025,
  title={{rLLM}: A Framework for Post-Training Language Agents},
  author={Tan, Sijun and Luo, Michael and Cai, Colin and Venkat, Tarun and Montgomery, Kyle and Hao, Aaron and Wu, Tianhao and Balyan, Arnav and Roongta, Manan and Wang, Chenguang and Li, Li Erran and Popa, Raluca Ada and Stoica, Ion},
  year={2025},
  note={Berkeley Sky Computing Lab},
  howpublished={\url{https://github.com/rllm-org/rllm}}
}

@article{agentlightning2025,
  title={Agent Lightning: Train ANY AI Agents with Reinforcement Learning},
  author={Xufang Luo and Yuge Zhang and Zhiyuan He and Zilong Wang and Siyun Zhao and Dongsheng Li and Luna K. Qiu and Yuqing Yang},
  journal={ArXiv},
  year={2025},
  volume={abs/2508.03680},
  url={https://api.semanticscholar.org/CorpusID:280526917}
}

\appendix
\section{Availability and License}
\label{sec:appendix}

MCP-U RL is released as open source under the Apache-2.0 license. The release includes
the three layers described in the paper, and
runnable configurations for the three domains of Section~\ref{sec:exp:capability}, so
every training curve reported here can be reproduced from a task specification and a
single launch command. Links to the code and project page are given with the author information on the first page.

\section{Training Details}
\label{sec:appendix:train}

All runs use the open-weight \texttt{gpt-oss-20b}~\cite{openai2025gptoss} policy
trained with GRPO~\cite{shao2024deepseekmath}. Shared hyperparameters are in
Table~\ref{tab:hparams}; settings that differ by domain are in
Table~\ref{tab:perdomain}.

\begin{table}[t]
  \centering
  \small
  \begin{tabular}{ll}
    \toprule
    Hyperparameter & Value \\
    \midrule
    Policy model            & \texttt{gpt-oss-20b} \\
    RL algorithm            & GRPO \\
    Learning rate           & $2\times10^{-6}$ \\
    Batch Size              & 16 \\
    Trajectories per task ($n$) & $8$ \\
    Reward                  & binary, from evaluator \\
    Optimizer               & AdamW \\
    KL coefficient          & 0.0 \\
    Rollout temperature     & 0.7 \\
    Max seq.\ len  & 128k \\
    \bottomrule
  \end{tabular}
  \caption{Hyperparameters shared across all three domains.}
  \label{tab:hparams}
\end{table}

\begin{table}[t]
  \centering
  \small
  \resizebox{\columnwidth}{!}{%
  \begin{tabular}{lccc}
    \toprule
    & Software eng. & Deep research & General tool use \\
    \midrule
    Dataset        & R2E-Gym & DeepDive & AgentWorldModel-1k \\
    Tasks    & 4574 & 2234 & 3315 \\
    Agent          & mini-swe-agent & ReAct & ReAct \\
    Max turns      & 100 & 40 & 40 \\
    Dataset link
      & \href{https://huggingface.co/datasets/R2E-Gym/R2E-Gym-Subset}{HF}
      & \href{https://huggingface.co/datasets/zai-org/DeepDive}{HF}
      & \href{https://huggingface.co/datasets/Snowflake/AgentWorldModel-1K}{HF} \\
    \bottomrule
  \end{tabular}}
  \caption{Training settings that differ by domain.}
  \label{tab:perdomain}
\end{table}

\section{Rollout Experiment Details}
\label{sec:appendix:rollout}

The throughput experiments of Section~\ref{sec:exp:system} run on the
software-engineering workload on 3$\times$H200 GPUs without high-speed NVLink
interconnect, so weights and trajectories move between GPUs over ordinary network
transfer rather than a fast GPU fabric. For the per-stage concurrency
sweep (Figure~\ref{fig:decouple}), the acquire stage is fixed at 24 workers and the
number of run workers is varied over $\{24, 48, 64, 96, 120\}$, with the in-flight
capacity (number of live environments) held fixed at 120 so the memory budget does
not change. Throughput is measured as mean rollout tokens per second over 5 steps after warm-up.

For the placement comparison (Table~\ref{tab:placement}), both placements run at
matched concurrency ($120$). The colocated placement shares GPUs between rollout and
the policy update and alternates between them; the fully asynchronous placement runs
rollout and the update on separate GPUs and streams trajectories through a queue. The
asynchronous placement is \emph{one-step off-policy}: the update is applied to
trajectories generated under a policy at most one step stale, which keeps it close to
on-policy while overlapping rollout and the update. Time per step is the mean over
the first 5 steps.

\section{Environments and MCP Servers}
\label{sec:appendix:env}

Each domain reaches its tools through one or more MCP servers, and the
environment-orchestration layer provisions them on either the Docker or the
daemonless backend depending on how environments are created
(Section~\ref{sec:design:env}).

\begin{itemize}
  \item \textbf{Software engineering.} Each task carries its own repository image, so
    environments are provisioned on demand on the daemonless (Apptainer) backend,
    built from the task's existing Dockerfile. The agent is given a single shell tool
    exposed over MCP; the hidden test suite runs inside the same container.
  \item \textbf{Deep research.} Every task is stateless and shares the same two MCP
    servers, a Google Search server (\texttt{serper-search}) and a page-reading server
    (\texttt{jina-scrape}), on the Docker backend. Because the configuration is shared,
    the pool is pre-warmed and environments are reset and reused across tasks.
  \item \textbf{General tool use.} Each AgentWorldModel environment is code- and
    database-backed and exposed over MCP, provisioned on demand on the daemonless
    (Apptainer) backend built from a Dockerfile.
\end{itemize}

\section{Evaluators}
\label{sec:appendix:eval}

Rewards come entirely from each task's evaluator, run against the environment after
the episode.

\begin{itemize}
  \item \textbf{Software engineering.} The evaluator runs the task's hidden test
    suite inside the container the agent edited and returns a binary reward on whether
    all tests pass.
  \item \textbf{Deep research.} The evaluator compares the agent's final answer to a
    reference answer with an LLM judge, returning a binary correctness reward.
  \item \textbf{General tool use.} Each task carries a deterministic SQL or code
    verifier that queries the application's database state after the episode and
    returns a binary reward on whether it matches the task's expected state.
\end{itemize}

\end{document}